# Healthcare Cost Prediction: Leveraging Fine-grain Temporal Patterns

Mohammad Amin Morid, PhD[1], Olivia R. Liu Sheng, PhD[2], Kensaku Kawamoto, MD, PhD, MHS[3], Travis Ault[4], Josette Dorius[4], Samir Abdelrahman, MS, PhD[3,5]

[1] Department of Information Systems and Analytics, Santa Clara University, CA, USA;

[2] Department of Operations and Information Systems, University of Utah, UT, USA;

[3] Department of Biomedical Informatics, University of Utah, UT, USA;

[4] University of Utah Health Plans, UT, USA;

[5] Computer Science Department, Cairo University, Egypt.

## ABSTRACT

*Objective*: To design and assess a method to leverage individuals' temporal data for predicting their healthcare cost. To achieve this goal, we first used patients' temporal data in their fine-grain form as opposed to coarse-grain form. Second, we devised novel spike detection features to extract temporal patterns that improve the performance of cost prediction. Third, we evaluated the effectiveness of different types of temporal features based on cost information, visit information and medical information for the prediction task.

*Materials and methods*: We used three years of medical and pharmacy claims data from 2013 to 2016 from a healthcare insurer, where the first two years were used to build the model to predict the costs in the third year. To prepare the data for modeling and prediction, the time series data of cost, visit and medical information were extracted in the form of fine-grain features (i.e., segmenting each time series into a sequence of consecutive windows and representing each window by various statistics such as sum). Then, temporal patterns of the time series were extracted and added to fine-grain features using a novel set of spike detection features (i.e., the fluctuation of data points). Gradient Boosting was applied on the final set of extracted features. Moreover, the contribution of each type of data (i.e., cost, visit and medical) was assessed. We benchmarked the proposed predictors against extant methods including those that used coarse-grain features which represent each time series with various statistics such as sum and the most recent portion of the values in the entire series. All prediction performances were measured in terms of Mean Absolute Percentage Error (MAPE).

*Results*: Gradient Boosting applied on fine-grain predictors outperformed coarse-grain predictors with a MAPE of 3.02 versus 8.14 (p<0.01). Enhancing the fine-grain features with the temporal pattern extraction features (i.e., spike detection features) further improved the MAPE to 2.04 (p<0.01). Removing cost, visit and medical status data resulted in MAPEs of 10.24, 2.22 and 2.07 respectively (p<0.01 for the first two comparisons and p=0.63 for the third comparison).

*Conclusions*: Leveraging fine-grain temporal patterns for healthcare cost prediction significantly improves prediction performance. Enhancing fine-grain features with extraction of temporal cost and visit patterns significantly improved the performance. However, medical features did not have a significant effect on prediction performance. Gradient Boosting outperformed all other prediction models.

**Keywords**: *healthcare cost prediction, temporal pattern extraction, temporal abstraction, machine learning*

## 1. INTRODUCTION

The United States' national health expenditure (NHE) grew 4.3% to $3.3 trillion in 2016 (i.e., $10,348 per person), which accounted for 17.9% of the nation's gross domestic product (GDP) [1]. In seeking to control and reduce these unsustainable increases in healthcare costs, it is imperative that healthcare organizations can predict the likely future costs of individuals, which can benefit various stakeholders. For health insurers and increasingly healthcare delivery systems, accurate forecasts of likely costs can help with business planning in different ways. First, care management resources can be efficiently targeted to those individuals at highest risk of incurring significant costs [2]. Moreover, due to the transition from fee-for-service payment models to value-based payment models [3], various healthcare organizations are actively seeking to reduce individuals' costs by providing different types of interventions (e.g., arranging frequent check-ups, home visits and telephone visits [4]). However, the effectiveness of such programs is difficult to assess since prospective randomized study designs are not routinely employed in these settings, making it difficult to know what individuals' costs would have been without these interventions. As a result, prediction of individuals' costs without the intervention may be the only available way to analyze return of investment and to enhance such programs in a data-driven manner.

Recently, we conducted a comprehensive literature review on healthcare cost prediction [5]. We found that healthcare cost prediction methods proposed to date are suboptimal. One reason for the suboptimal performance is that many approaches do not use individuals' cost history to predict future cost. A second reason is that the few studies that used cost predictors generally limited their analyses to coarse-grain features with high information loss. More specifically, they generally summarized individuals' temporal

(time series) data by a set of abstracted values (e.g., overall costs over the past year, medical costs over the last six months, pharmacy costs over the last three months). This problem exists for non-cost predictors as well. Finally, we found no studies that accounted for the effect of temporal fluctuations (e.g., spikes) on future costs.

In this study, we aimed to develop an advanced feature engineering method to predict healthcare costs by addressing the above limitations and improving their performance in two ways. First, the proposed method benefits from cost, visit and medical predictors at a fine-grain level, as opposed to a coarse-grain level. Second, temporal patterns of individuals' behavior were captured by using a proposed set of spike detection features and fed into the prediction model. Moreover, we assessed the relative importance of cost, visit and medical data for cost prediction. More specifically, this study answered the following research questions:

"What is the effect of cost, visit and medical features on healthcare cost prediction performance?"

"How should we leverage fine-grain time series features of individuals' cost, visit and medical status in order to improve predictive performance?"

"What is an effective approach to detect temporal patterns in individuals' data to improve predictive performance?"

## 2. COST PREDICTION LITERATURE REVIEW

Our literature review [5] showed that prior work on the use of supervised learning for cost prediction were of three types. In the first type, the research goal is to predict cost using medical predictors and to show the effect of medical factors such as chronic disease scores on cost prediction [6]. In the second type of study, cost predictors with or without non-cost predictors (e.g., medical or visit data) are used to predict cost. In the last type of study, researchers bucket individuals' costs and predict the cost bucket rather than the actual cost value. We followed the second general approach, since we aimed at predicting cost using cost, visit and medical status predictors.

In our literature review[5], we identified five studies that used cost to predict future healthcare costs [2, 7-10]. Tables 1 and 2 summarize the different types of inputs, prediction models and outputs of these five studies. Among these studies, Bertsimas et al. (2008) [7] used fine-grain predictors to a limited extent by using cost from each of the last 12 months in their prediction process. However, the rest of the studies used coarse-grain predictors by using inputs such as total pharmacy cost and total medical cost. Moreover, Bertsimas et al. (2008) [7] was the only study to involve temporal patterns in the prediction process by detecting spikes (i.e., fluctuation points) in patients' temporal data using *number of months*

*above average*, *cost of the highest month* and a variable they termed *acute* (which indicates if the highest month's cost is significantly different than the average of monthly costs). Also, they found that adding medical and demographic features to their cost predictors did not improve the prediction performance. Finally, while different medical features have been evaluated for cost prediction, there is just one study by Duncan et al. (2016) [2] that used visit information, and this study simply used the total number of visits. The features proposed by Bertsimas et al. (2008) [7] were used as a baseline set of features in our experiments, since they were the most complete set of features we found in the literature.

Table 1 - Input features used for published methods to predict future cost based on prior cost. The numbers between parentheses in all columns (except the first column) refer to the number of variables (i.e., inputs).

| Paper | Number of Types of Cost Inputs | Cost Inputs | Non Cost Inputs |
| --- | --- | --- | --- |
| Bertsimas (2008)[7] | 21 | Monthly cost (12), Total pharmacy cost, Total medical cost, Total cost, Total cost in last 6 months, Total cost in last 3 months, Trend, Acute, Months above average, Cost of highest month | Age, Gender, Sex, Diagnosis code groups (218), Procedure groups (180), Drug groups (336) |
| Duncan (2016)[2] | 4 | Professional costs, Pharmacy costs, Outpatient costs, Inpatient costs | Age, Sex, Diagnosis code groups (83), Total visits, Primary care provider visits |
| Sushmita (2015)[8] | 1 | Total previous cost | Age, Sex, Diagnosis code groups (211), Procedure groups (233), Comorbidity scores |
| Kuo (2011)[10] | 1 | Previous medication cost | Age, Sex, Elixhauser's index, Pharmacy-based metrics |
| Frees (2013)[9] | 1 | Total previous cost | Sex, Race, Region, Education, Job, Marriage, Income level, Self-rated physical health, Self-rated mental health |

Table 2 - Regression models used for published methods to predict future cost based on prior cost.

| Paper | Method | Outcome |
|---|---|---|
| Duncan (2016)[2] | Gradient Boosting Decision Tree, Lasso, M5 | Paid amount |
| Sushmita (2015)[8] | M5, Random Forest, CART | Billed amount |
| Frees (2013)[9] | Linear regression | Paid amount |
| Kuo (2011)[10] | Linear regression | Billed amount |
| Bertsimas (2008)[7] | CART | Paid amount |

This study aimed at benefiting from fine-grain temporal abstraction in an extended way as well as proposing a set of innovative spike detection features to leverage temporal patterns.

## 3. BACKGROUND

### 3.1 Temporal abstraction

Temporal abstraction transforms time series data into the input features of a classification model [11]. One commonly used temporal abstraction approach is to segment a time series into a sequence of fixed-sized non-overlapping consecutive windows or intervals [12]. Then, each window is represented by an aggregation measure (e.g., average, sum, count) of all data values time-stamped within the window. We refer to the window size as the grain size and divide the temporal features into fine-grain (i.e., small window size) features versus coarse-grain (i.e., large window size) features.

In the healthcare cost prediction literature, features such as prior medication cost or total cost in the last six months are considered coarse-grain temporal abstraction features, while monthly costs over the last 12 months are considered fine-grain temporal abstraction features. To our knowledge, our study is the first study to analyze the effect of temporal abstraction on healthcare cost prediction.

### 3.2 Temporal Pattern Detection

This study aimed at leveraging patients' temporal data for healthcare cost prediction by benefiting from features extracted from their temporal patterns. To achieve this, we leveraged pattern detection methods that have been widely used for tasks such as image recognition, speech analysis, traffic analysis and smog detection [13]. This section gives a brief overview of these pattern detection methods.

The aim of pattern detection is to identify an object (e.g., patient) as belonging to a particular group (e.g., high cost or low cost) by extracting patterns and regularities that are specific to that group [14]. The underlying idea is that the objects associated with a particular group share common attributes (i.e., patterns) more than the objects in other groups [15]. Pattern recognition can be divided into two basic tasks: description and classification [16]. The first task extracts features from each object using feature extraction techniques to represent that object. The second task assigns a group label to the object based on the extracted attributes using a classification method.

Temporal pattern detection can be implemented using various methods. Change point detection is one of the most popular approaches for this purpose. A wide range of disciplines have utilized a variety of change points in temporal, spatial and other types of data sequences to help with quality control, robot control, and fraud detection applications amongst others [17, 18]. Past healthcare research also has emphasized the importance of changes in patients' health signals in clinical guidelines and decision support [19-21]. Hence, we hypothesized that creating and utilizing features based on changes in a patient's time series would improve the accuracy of predicting patients' health outcomes such as cost.

A change point is a point in time, location or another type of sequence index at which a change or a fluctuation occurs in a data sequence. The definitions of change points vary by disciplines and studies. Zhou et al. [17] introduced four categories of definitions that define change points based on changes in statistical parameters (e.g. mean and/or variance of a statistical distribution), actual values, models fitted to data and derived attributes (e.g., predicted values). Spike detection, which is the focus of this study, is also considered as a change point detection method [22]. In our study, we used the pruned exact linear time (PELT) [23] method, which is capable of detecting multiple change points (i.e., spikes) with respect to changes in mean regardless of the statistical distribution of the data. PELT splits time series data into multiple segments where the mean value of each segment is significantly different from its prior and posterior segments. A list of the first values of each change point segment (CPS) is the output of the PELT method. The main advantage of PELT over other change point detection methods is that its computational cost is linear to the length of the input time series.

## 4. METHOD

In this research, the first goal was to study cost prediction using fine-grain versus coarse-grain model inputs. The second goal was to leverage the fine-grain data with our proposed spike detection method to extract temporal features. The third goal was to assess the effect of different type of predictors (i.e., cost, visit and medical status) on future costs. In the rest of this section, we describe each goal in further detail. More specifically, Section 4.1 describes how we extracted cost, medical and visit features from medical and pharmacy claims data. Section 4.2 explains the temporal abstraction method used for representing the time series of the extracted features. Finally, Section 4.3 elaborates on our proposed spike detection method.

**4.1 Cost, medical and visit predictors**

Although there has been past research on the effect of cost and non-cost predictors on individuals' future costs, we found no studies that evaluated various features proposed in different literature on a single benchmark to compare their effect against each other. This study used cost predictors as well as non-cost predictors to analyze the impact of each predictor on future cost prediction.

Since different sets of medical predictors have been proposed in different studies, we implemented all medical predictors included in prior identified studies (shown in Table 1) to find the best set for our evaluation. The preliminary analyses showed that the 180 procedure groups and the 336 drug groups used by Bertsimas et al. (2008) [7] and the 83 diagnosis groups used by Duncan et al. (2016) [2] have the best performance. Therefore, these sets of variables were extracted to represent patients' medical information (see Table 3). These variables are continuous representing the number of claims for a member that have coding belonging to the different procedure, diagnosis and drug groups.

While there has been very limited study on the effect of patient visit information on their cost, such visit information could be very useful. Two patients with the same amount of cost in a specific month may have a different visit pattern. For example, patients who are constantly frequent visitors may be at a higher risk of high future costs compared to those who are temporarily frequent visitors.

As mentioned before, visit information of patients has not been studied well for cost prediction. In this paper, we proposed to use seven categories of visits for the cost prediction task including office, inpatient, outpatient, lab, emergency, home and others. All cost predictors as well as non-cost predictors are summarized in Table 3.

Table 3 - Time series features used as inputs for this study.

| Category | Time Series Features | Cost Inputs | Reference |
|---|---|---|---|
| Cost | 2 | Medical cost and pharmacy cost | |
| Medical | 180 | Procedure groups | Bertsimas (2008)[7] |
| Medical | 83 | Diagnosis code groups | Duncan (2016)[2] |
| Medical | 336 | Drug groups | Bertsimas (2008)[7] |
| Visit | 7 | Office, Inpatient, Outpatient, Lab, Emergency, Home, Other | |

## 4.2 Fine-grain and coarse-grain temporal abstraction

Fine-grain abstraction leads to high dimensionality, but low information loss, while coarse-grain abstraction results in low dimensionality, but high information loss. Our hypothesis was that benefiting from fine-grain abstraction features could afford improved accuracy over coarse-grain features for healthcare cost prediction.

To implement fine-grain temporal abstraction, piecewise aggregation approximation (PAA) [12] was used to represent the monthly data of each individual. More specifically, we divided the time series of each input feature (i.e., cost, visit and medical features) into fixed sized windows. Then, the total amount of the individual's costs in each cost group (e.g., medical and pharmacy), the total number of the individual's visits in each visit group (e.g., inpatient and outpatient) and the total number of the individual's claims in each medical group (e.g., procedure, diagnosis and drug) were used to represent the values in each time window. Here we examined the following hypothesis:

*H1: Fine-grain predictors are more effective than coarse-grain predictors.*

After trying different window sizes for fine-grain segmentation of data, the 1-month window size resulted in the best performance. Table s1 in the online supplement contains the experimental results. Therefore, since two years of patients' data were used to predict their cost in the following year, each input time series was represented by 24 data values. These time series features are shown in Table 3.

## 4.3 Temporal pattern detection

Ignoring temporal patterns in individuals' data can result in misclassifying individuals to the same category despite different trends in their temporal values. For example, Figure 1 shows the time series of two patients that have a similar amount of total cost but different temporal patterns. As seen, although the total healthcare costs for these two patients are very similar, they have incurred these costs in very different ways. Patient B has constantly been a high cost member and shows a typical pattern for a member with a chronic condition. Such a constant high cost pattern has a strong tendency to repeat in the future. On the other hand, Patient A is considered as a low cost member, except month 17 when the patient has a spike due to an exceptional situation (e.g., pregnancy or accident). Such a cost pattern that exhibits a spike might have a low risk of high future costs.

These types of constant or spike patterns are not just limited to individuals' costs. Patients with a similar number of visits may have a very different visit pattern for the same reason explained above. This story can be the same for all other aspects of patients. These spikes are the fluctuation points in time series that are the focus of change detection to extract temporal patterns.

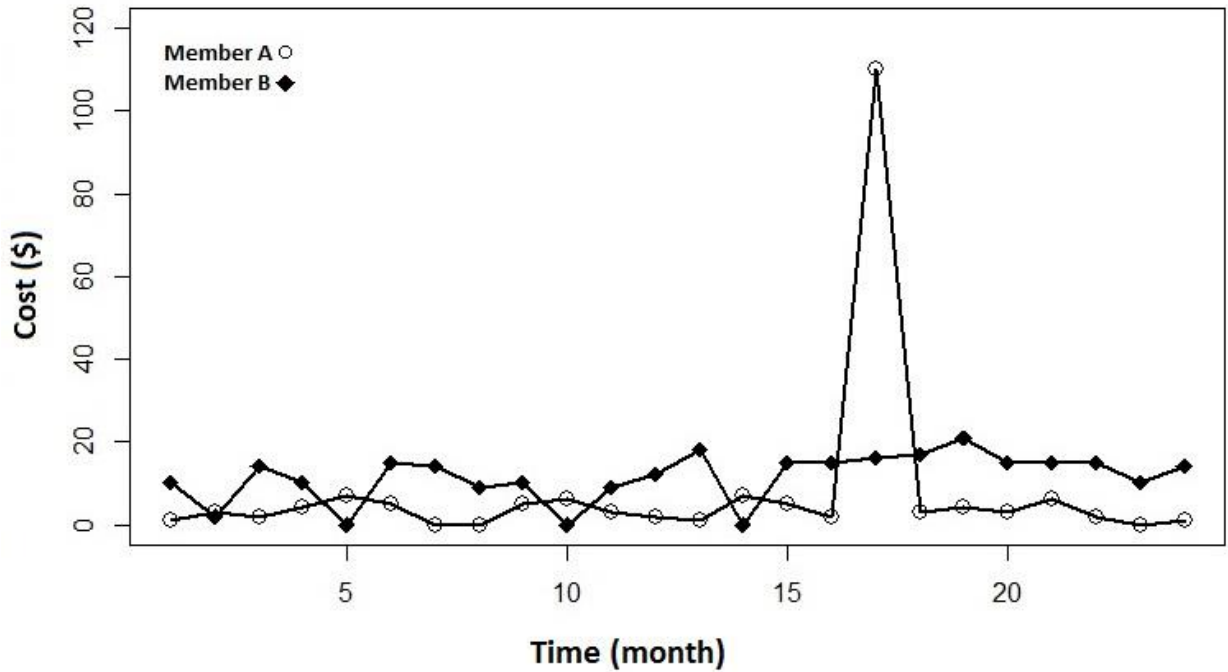

Figure 1 - Total healthcare costs of two patients with the same overall cost.

Although Bertsimas et al. (2008) [7] was the first and only study we found that incorporated this consideration in their cost prediction model, their proposed approach for leveraging such cost difference patterns was limited. Specifically, there were three features that they proposed in their study to represent the spikes: (i) *acute* (whether or not the highest month cost is significantly different from the average), (ii) *cost of the highest month* and (iii) *number of months above average*. The first two features are based on having a single spike in a patient's cost profile, which may not always be the case. Also, although the third feature attempts to catch more than one spike, it can weaken prediction model performance as illustrated by the following example (Figure 2).

As seen in Figure 2, most cost data points are above the average, while there are very few spikes in the profile. Moreover, although the patient has spikes in his or her profile, they are not remarkable in terms of the value, since all the cost values are relatively low.

In our study, we detected spikes by using change point detection methods. This was done in four steps: 1) each time series (shown in Table 3) is segmented into 24 windows (as described in the previous section); 2) the fine-grain value of each window is extracted for each window; 3) change points are detected using PELT [23]; and 4) temporal patterns are detected using a proposed set of features. The proposed method is described in more detail below.

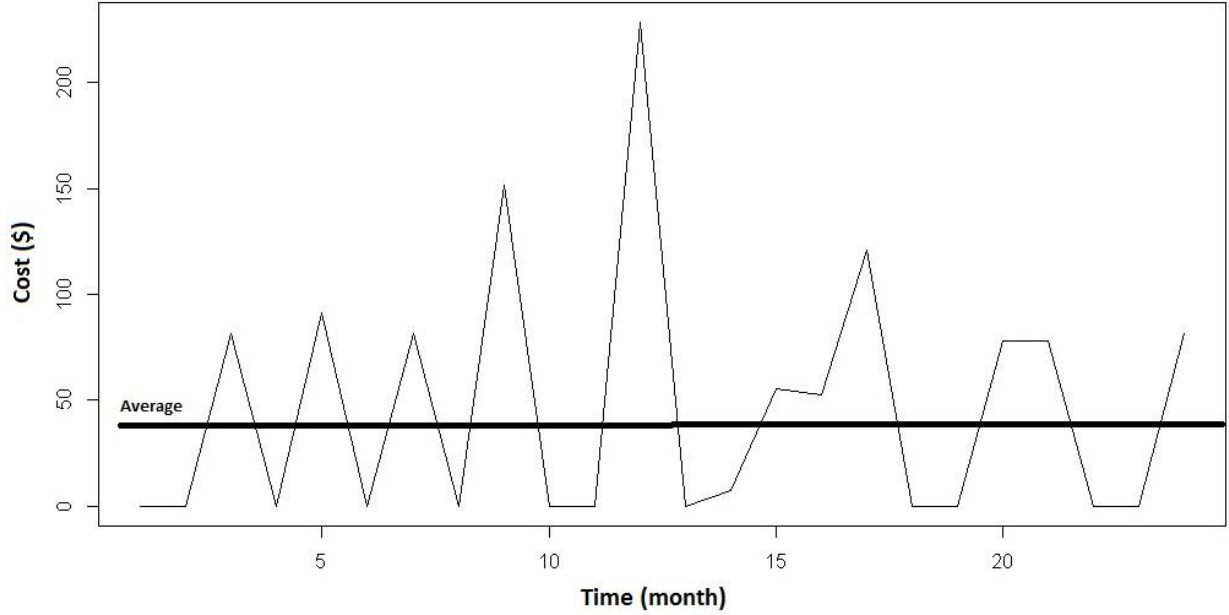

Figure 2 - An example cost profile pattern where most data points are above the average.

We denote our time series as $t_{1:n} = (t_1,...,t_n)$, where $t$ is the time series of fine grained values of $n$ windows (e.g., $n$ is 24 in this study). The PELT output gives $m+1$ segments and $m$ change points as $\tau_{1:m} = (\tau_1,...,\tau_m)$, where the change points are ordered such that $\tau_k < \tau_l$ if, and only if, $k < l$ (i.e., $k$ and $l$ are two consecutive change points), and $t_{(\tau_{k-1}+1)}: t_{\tau_k}$ denotes one change point segment [23]. The goal is to find a set of change points where the statistical properties of each CPS is different from its previous and posterior CPSs. PELT identifies multiple change points by minimizing the following:

$$\sum_{k=1}^{m+1}[C(t_{(\tau_{k-1}+1)}: t_{\tau_k})] + \rho f(m) \tag{1}$$

Here $C$ is a cost function for a segment and $\rho f(m)$ is a penalty to avoid overfitting. Aligned with Rigaill (2010) [24] we use quadratic loss as the cost function (i.e., $C$) for this study. The most common choice for penalty function is $\rho f(m) = \rho m$ which penalizes overfitting with a linear relationship to the number of change points [25]. Trying different values for $\rho$, our preliminary analysis showed that $\rho=2$ had the best performance, and so it was selected as the final choice. PELT avoids trying all possible values for $m$ and just checks a promising set of possible values. To do so, the algorithm minimizes equation (1) using dynamic programming and pruning techniques to reach the optimal segmentation. The computational time reduction is mainly achieved by the assumption that the number of change points increases linearly as the data set grows.

Having the change point values detected by PELT, let *State(i,f,τ$_j$)* denote the *state* of patient *i* during change point segment (CPS) *j* for feature *f*. For example, this *state* for the cost feature denotes a patient's total cost amount for a specific time window, while for inpatient visits the *state* of this feature denotes the number of inpatient visits. We define a change in the *state* of patient *i* at change point segment *j* for feature *f* as follows:

*Change(i,f,j)*   = *Increase (I), if State(i,f,τ$_j$) > State(i,f,τ$_{j-1}$)*                                                           (2)

             = *Decrease (D), if State(i,f,τ$_j$) < State(i,f,τ$_{j-1}$)*

In time series research, temporal patterns can be derived based on gradient values of a numeric series [16, 26]. Our study focuses on sharp, discrete gradient changes in patients' state, since sharp changes in a patient's status level (e.g., cost, visit) might be temporary. For instance, this can differentiate temporally high cost patients from those who are permanently high cost. We extract this type of temporal change pattern by detecting spikes in a patient's state based on a tandem pattern of an *Increase Change* immediately before a *Decrease Change* in a time series as follows:

$Spike(i,f,j) = 1$, *if Change(i,f,j)= I and Change(i,f,j+1)=D,  for j = 1,...m-1*                      (3)

where Spike(i,f,j) for all change point segments were initialized to zero. Finally, the total number of *Spikes* is counted in this study as follows:

$$Count\_of\_spike(i,f) = \sum_{k=1}^{m} Spike(i,f,k) \qquad (4)$$

Different from the spike features in Bertsimas et al. (2008) [7], we proposed an advanced spike detection method which is capable of detecting multiple spikes. Moreover, *Spike* may have a different meaning for different patients for cost prediction. For instance, for a low cost patient who has a few medical claims in a year with $50 in cost on average, $300 is considered a spike. However, even though these are considered as *Spikes* by change point detection methods, healthcare providers may not be worried about such kinds of *Spikes*. Therefore, besides the number of *Spikes*, the amount of positive and negative changes before and after that is important.

To calculate the amount of positive and negative changes, first, we define the amount of change as follows:

$$ChangeAmount(i,f,j) \quad = State(i,f,\tau_j) - State(i,f,\tau_{j-1}), \text{ if } State(i,f,\tau_j) > State(i,f,\tau_{j-1}) \quad (5)$$

$$= State(i,f,\tau_{j-1}) - State(i,f,\tau_j), \text{ if } State(i,f,\tau_j) < State(i,f,\tau_{j-1})$$

Then, the positive and negative amounts of changes are calculated as follows:

$$Amount\_of\_positive\_changes(i,f) = \sum_{k=1}^{m} ChangeAmount(i,f,k), if\ Spike(i,f,k) = 1 \quad (6)$$

$$Amount\_of\_negative\_changes(i,f) = \sum_{k=1}^{m-1} ChangeAmount(i,f,k+1), if\ Spike(i,f,k) = 1 \quad (7)$$

For example, Figure 3 shows an example of the proposed spike detection features for the same data as Figure 2. Each circle shows a change point and each bold line shows a change point segment (CPS) as a result of applying PELT on the time series of 24 data values extracted from fine-grain abstraction using the PAA approach described in the previous section. Out of all change points, spikes are those that occurred on the $81, $91, $228 and $85 increases (i.e., positive changes). The amount of positive changes and the amount of negative changes are calculated by adding the increase (i.e., positive change) and the decrease (i.e., negative change) in amounts according to Eq. 6 and Eq. 7 respectively.

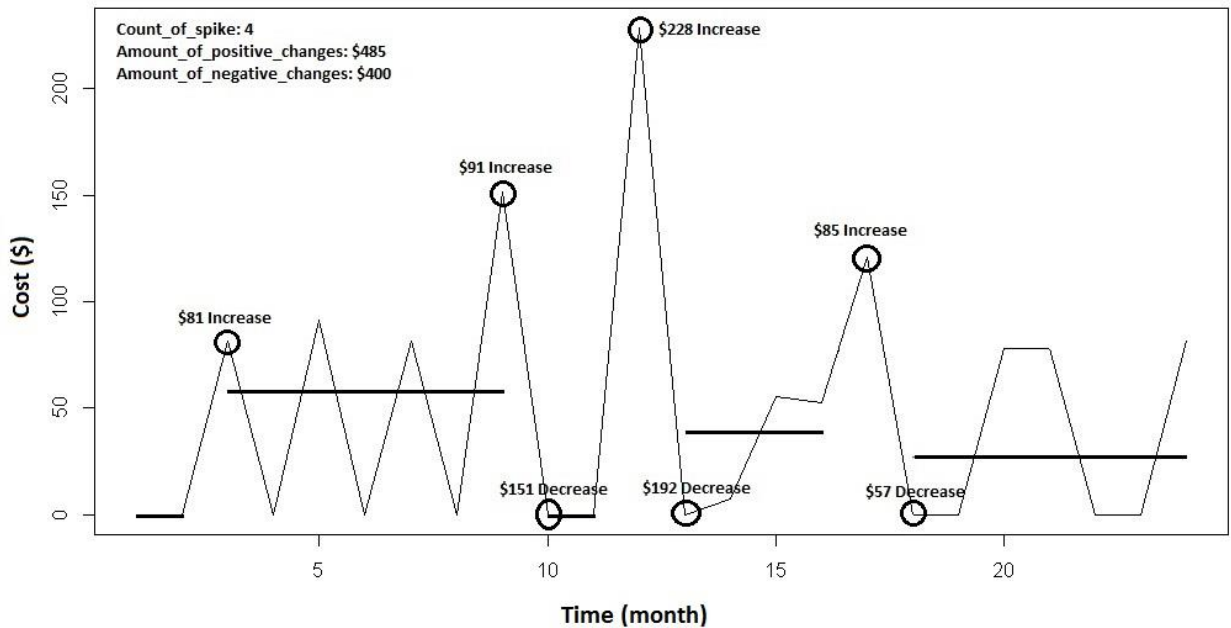

Figure 3 - An example of extracted spike detection features proposed in this study.

Our algorithm to generate the spike detection features is summarized in algorithm 1 as follows:

Algorithm 1 – Spike detection

*Input*: Raw time series of F features for P patients *(i.e., $d_{1:r,i,f}$)*

*Output:* Spike detection features for the time series *(i.e., Count_of_spike(i,f), Amount_of_positive_changes(i,f)* and *Amount_of_negative_changes(i,f))*

*for patient i in (1..P):*

    *for feature f in (1..F):*

        *$t_{1:n,i,f,}$ = PAA ($d_{1:r,i,f}$)*

        *State(i,f) = PELT($t_{1:n,i,f}$)*

        *Initialize Change(i,f) with Null (N)*

        *Initialize ChangeAmount(i,f) with 0*

        *for change point j in (2..m):*

            *if State(i,f,$\tau_j$) > State(i,f,$\tau_{j-1}$):*

                *Change(i,f,j) = Increase (I)*

                *ChangeAmount(i,f,j) = State(i,f,$\tau_j$) - State(i,f,$\tau_{j-1}$)*

            *else if State(i,f,$\tau_j$) < State(i,f,$\tau_{j-1}$):*

                *Change(i,f,j) = Decrease (D)*

                *ChangeAmount(i,f,j) = State(i,f,$\tau_{j-1}$) - State(i,f,$\tau_j$)*

        *for change point j in (2..m):*

            *if Change(i,f,j-1)= I and Change(i,f,j)=D:*

                *Spike(i,f,j) = 1*

                *Amount_of_positive_changes(i,f) += ChangeAmount(i,f,j)*

                *Count_of_spike(i,f) += 1*

            *else if Change(i,f,j-1)= D and Change(i,f,j)=I:*

                *Spike(i,f,j) = 0*

                *Amount_of_negative_changes(i,f) += ChangeAmount(i,f,j)*

Therefore, our underlying hypothesis for the proposed spike detection features is that they are an effective way of detecting temporal patterns for cost prediction. Here we examined the following hypothesis:

*H2: Detecting temporal patterns using the number of spikes and the actual amount of changes improves the performance of cost prediction.*

## 5. EMPIRICAL EVALUATION

In the evaluation step, we conducted four experiments to evaluate the effectiveness of medical, visit and cost predictors, test the two aforementioned hypotheses, and analyze the importance of each proposed change point features. In each experiment we extracted a set of features from the 608 input time series (i.e., 2 cost, 180 procedure, 83 diagnosis code, 336 drug and 7 visit features) described in Table 3. The extracted features (i.e., predictors) are described in Table 4. The target cost variable in this study is the paid amount which is the amount of money paid by a healthcare insurer to its members.

The first experiment attempted to show the effect of different predictors including cost, visit and medical information on future cost prediction. To achieve this, our numeric prediction approach was applied on the features proposed by Bertsimas et al. (2008) [7], as a baseline set of features. These features are the most complete set of features we found in the literature. They are a combination of fine-grain, coarse-grain and spike detection features as noted in the Method, shown by "C/F" in this study (see Table 4). This feature set was found to have strong prediction performance in our previous study[5]. Although this feature set was just proposed for cost predictors, we followed the same extraction approach for visit and medical predictors as well. Table 4 shows the extraction process of these features. For instance, *Overall_value* – the overall amount/count of values in the observation period – is a coarse-grain feature calculated for cost predictors. For visit predictors, the overall counts of visits for each visit group are calculated. For medical predictors, the overall counts of claims for each diagnosis or procedure or drug groups are calculated.

In the second experiment, the proposed fine-grain and coarse-grain features were introduced to our numeric prediction model to test H1 by assessing the effect of each type of temporal abstraction. For fine-grain features we used the features introduced in the Method and for the coarse-grain features we used the coarse-grain features of the baseline feature set. These features are shown by "F" and "C" in the experiments respectively (see Table 4).

In the third experiment, testing H2, our goal was to assess the effect of the proposed spike detection features (i.e., *Count_of_spike*, *Amount_of_positive_changes*, *Amount_of_negative_changes*) (see Table 4). To achieve this, the performance of the proposed spike detection features combined with the proposed fine-grain features, referred to as "FS", was compared against the baseline features (i.e., "C/F"), the coarse-grain features (i.e., "C") and the proposed fine-grain features (i.e., "F"). Although this study was on predicting the actual costs, we attempted to compare the performance of different numeric predictors according to their classification as well. This approach can give a better sense of the importance of the numeric prediction models on different cost buckets. To achieve this, the target variable was changed to predicting a patient's cost in the future year in terms of different cost buckets.

Table 4 – Extracted input features from each time series described in Table 3.

| Name | Description | Type | C* | C/F* | F* | FS* |
|---|---|---|---|---|---|---|
| Overall_value | Overall amount/count of values in the observation period | Coarse grain | * | * | | |
| Six_months_value | Overall amount/count of values in the last 6 months of the observation period | Coarse grain | * | * | | |
| Three_months_value | Overall amount/count of values in the last 3 months of the observation period | Coarse grain | * | * | | |
| Trend | Slope of last year's monthly data in the observation period | Coarse grain | * | * | | |
| Acute | If the amount/count of the highest month value is significantly different than the average, the indicator takes on the value 1. | Change point | | * | | |
| Highest_value | The amount/count of the highest month in the observation period | Change point | | * | | |
| Num_above_average | Number of months with amount/count values above the average. | Change point | | * | | |
| Last_year_monthly_value | Monthly amount/count of values in the last twelve months of the observation period | Fine grain | | * | | |
| Proposed fine grain features | Monthly amount/count of values in the twenty four months of the observation period | Fine grain | | | * | * |
| Proposed spike detection features | Count_of_spike, Amount_of_positive_changes, Amount_of_negative_changes | Spike detection | | | | * |

*C: Most commonly used coarse-grain features in the literature. C/F: Most commonly used set of coarse-grain, fine-grain, and spike detection features in the literature. F: proposed set of fine grain features. F/S: proposed set of fine-grain and spike detection features. Same designations are used in subsequent tables.

Finally, shedding more light on the effect of each proposed spike detection feature, in the fourth experiment we removed one of these features (i.e., *Count_of_spike*, *Amount_of_positive_changes*, *Amount_of_negative_changes*) (see Table 4) at a time and evaluated the performance of the rest of the spike detection features.

**5.1 Data**

Our data set consisted of 6.3 million medical claims and 1.2 million pharmacy claims from approximately 91,000 distinct individuals covered by University of Utah Health Plans from October 2013 to October 2016. Available data included demographic information (e.g., age, gender, age), clinical visit information (e.g., place and date of service, provider information), diagnosis and procedure codes, pharmacy dispense information, and cost information (e.g., paid, allowed and billed amount). This data were filtered to individuals with insurance membership for the whole three years period, which resulted in approximately 3.8 million medical claims and 780,000 pharmacy claims from 24,000 patients.

The data set was divided into two time periods: an observation period and a result period. The former time period was from October 2013 to September 2015 (i.e., two years), which was used to predict individuals' cost in the result period ranging from October 2015 to October 2016 (i.e., one year). Table 3 shows all input features used in this study and Table 5 shows demographic and insurance profile of the patients.

Table 5 – Demographic and insurance profile of patients included in analysis.

|  | Number of members | % of members |
|---|---|---|
| **Age** | | |
| 0-20 | 57765 | 63% |
| 20-40 | 18192 | 20% |
| 40-60 | 9489 | 10% |
| 60-80 | 4446 | 5% |
| 80-100 | 1381 | 2% |
| **Gender** | | |
| Female | 50931 | 56% |
| Male | 40343 | 44% |
| **Primary Insurance Provider** | | |
| Yes | 74311 | 81% |
| No | 16963 | 19% |
| **Insurance Type** | | |
| Medicaid | 87623 | 96% |
| Commercial | 3651 | 4% |

The range of paid amounts in the result period showed that 80% of the overall cost of the population came from only 15% of the members. Therefore, aligned with the literature on cost bucketing [7], to reduce the effects of extremely expensive members, the data set was partitioned into five different cost buckets where buckets 1 and 5 correspond to the lowest and highest cost buckets. This partitioning was done so that the sum of members' costs in each bucket was approximately the same in the observation period (i.e., the total dollar amount in each bucket was the same).

## 5.2 Evaluation Setup

For the numeric prediction task, several state-of-art models were applied on the extracted predictors in Table 4 to predict cost. To achieve this, each model was tuned to find its best performance. These models include including Linear Regression, Lasso [27], Ridge [28], Elastic Net [29], CART [30], M5 [31], Random Forest [32], Bagging [33], Gradient Boosting [33], and Artificial Neural Network [34]. Aligned with our previous study [5] as well as other studies [2], Gradient Boosting outperformed all other models and was chosen for this study. The performance results comparison of these models can be found in Table s2 of the online supplement. Since the focus of this paper is on finding the optimal input features for healthcare cost prediction, detailed result of the numeric prediction methods are not reported. Detailed conceptual comparison among numeric cost prediction methods can be found in our previous study [5]. 30% of the data set was used for the model selection and parameter tuning.

Cross validation was employed as the evaluation method on 70% of the data set. More specifically, the regression performance was measured according to the average of mean absolute percentage error (MAPE) across 20-folds. MAPE is the most common relative error measure for cost prediction used in the literature [5], and it was calculated as follows:

$$\frac{1}{N} \sum_{i=1}^{N} \left| \frac{Actual_i - Predicted_i}{Actual_i} \right| \qquad (8)$$

Here, $N$ is the number of instances (i.e., patients). To avoid division by zero, one dollar was added to all patients costs. Due to sensitive nature of the data, we are not able to report error measures that use absolute values (e.g., mean absolute error).

MAPE was evaluated across the 20 folds using pair-wise t-tests. Because of the large number of comparisons, Bonferroni correction [35] was performed as a post hoc test. Only p-values less than 0.01 were considered to be statistically significant at an alpha = 0.05. This statistical approach was aligned with the method recommended by Demsar [36]. For the third experiment, the classification performance was assessed in terms of accuracy, recall (i.e., hit ratio) and precision, with accuracy used as the primary measure for the t-tests. To further evaluate the proposed method, we also evaluated its performance according to a domain knowledge based measure, called Penalty Error [7]. This measure penalizes models for underestimating high cost members or overestimating low cost members. Table 6 shows the penalty table for the five-cost-bucket scheme used in this paper. The final value of the penalty error is calculated from the average forecast penalty per member of a given sample.

Table 6 - Penalty table based on the predicted and actual cost buckets.

|  | \multicolumn{6}{c|}{Actual Bucket} |
| --- | --- | --- | --- | --- | --- | --- |
|  | Bucket | 1 | 2 | 3 | 4 | 5 |
| **Predicted Bucket** | 1 | 0 | 1 | 2 | 3 | 4 |
|  | 2 | 1 | 0 | 1 | 2 | 3 |
|  | 3 | 2 | 1 | 0 | 1 | 2 |
|  | 4 | 3 | 2 | 1 | 0 | 1 |
|  | 5 | 4 | 3 | 2 | 1 | 0 |

### 5.3 Results

### 5.3.1 Performance of cost, medical and visit predictors

The MAPE performance of cost predictors for cost prediction were significantly higher than medical and visit predictors (2.8 versus 17.79 and 20.41; $p<0.01$). While adding visit predictors to the cost predictors improved the MAPE performance (2.61 versus 2.8; $p<0.01$), medical predictors did not have a significant effect (2.81 versus 2.8; $p=0.63$) (Table 7).

Table 7 - MAPE of the cost, visit and medical predictors as well as their combination for cost prediction over the five cost buckets.

| Predictors | All | 1 | 2 | 3 | 4 | 5 |
| --- | --- | --- | --- | --- | --- | --- |
| Cost + Visit | **2.61** | **2.42** | **3.1** | **3.85** | **5.54** | **7.45** |
| Cost + Medical | 2.81 | 2.58 | 3.25 | 4.37 | 7.01 | 9.14 |
| Cost | 2.8 | 2.57 | 3.22 | 4.32 | 6.94 | 9.11 |
| Medical | 17.79 | 17.64 | 18.06 | 19.25 | 19.99 | 20.83 |
| Visit | 20.41 | 20.11 | 21.81 | 22.06 | 23.54 | 24.31 |

### 5.3.2 Fine-grain and coarse-grain temporal abstraction

Similar to the first experiment, the combination of cost and visit predictors had the best MAPE performance both for the proposed fine-grain features as well as coarse-grain features ($p<0.01$ in all comparisons). The best MAPE performance of the proposed fine-grain features for cost prediction was significantly higher than coarse-grain features (3.02 versus 8.32; $p<0.01$) (Table 8).

Table 8 - MAPE of the proposed fine-grain features and coarse-grain features over different types of predictors for cost prediction over the five cost buckets.

| Feature | Cost Predictor | Visit Predictor | Medical Predictor | All | 1 | 2 | 3 | 4 | 5 |
|---|---|---|---|---|---|---|---|---|---|
| C |   |   | * | 20.61 | 20.14 | 22.08 | 23.66 | 27.69 | 29.17 |
| C |   | * |   | 23.18 | 22.71 | 24.56 | 26.78 | 29.19 | 32.14 |
| C | * |   |   | 8.76 | 8.35 | 9.09 | 12.43 | 16.96 | 19.12 |
| C | * | * |   | **8.32** | **7.94** | **8.75** | **11.66** | **15.29** | **17.11** |
| C | * |   | * | 8.78 | 8.37 | 9.1 | 12.45 | 16.96 | 19.14 |
| C | * | * | * | 8.55 | 8.17 | 8.89 | 11.93 | 15.84 | 17.78 |
| F | * |   |   | 3.27 | 2.76 | 5.73 | 6.16 | 7.99 | 8.92 |
| F |   |   | * | 15.65 | 15.27 | 16.29 | 18.97 | 21.94 | 23.48 |
| F |   | * |   | 17.21 | 18.94 | 20.78 | 22.64 | 24.51 | 28.15 |
| F | * | * |   | **3.02** | **2.73** | **4.16** | **4.74** | **6.95** | **7.87** |
| F | * |   | * | 3.29 | 2.78 | 5.78 | 6.18 | 8.13 | 9.02 |
| F | * | * | * | 3.04 | 2.74 | 4.17 | 4.78 | 6.98 | 7.94 |

### 5.3.3 Temporal pattern detection

The MAPE performance of the proposed spike detection features combined with fine-grain features for cost prediction was significantly higher than the baseline features, the fine-grain features and the coarse- grain features (2.02 versus 2.61, 8.32 and 3.02; $p<0.01$) (Table 9). Also, the MAPE performance of other numeric prediction methods applied on the proposed spike detection features combined with fine-grain features for cost prediction is provided in Table s2 of the online supplement. This table shows the superiority of the Gradient Boosting model.

Table 9 - MAPE of the proposed spike detection features combined with fine-grain features, the baseline features, the fine-grain features and the coarse-grain features for cost prediction over the five cost buckets.

| Feature | Cost Predictor | Visit Predictor | All | 1 | 2 | 3 | 4 | 5 |
|---|---|---|---|---|---|---|---|---|
| FS | * | * | **2.02** | **1.94** | **2.25** | **2.53** | **3.07** | **4.21** |
| C/F | * | * | 2.61 | 2.42 | 3.10 | 3.85 | 5.54 | 7.45 |
| F | * | * | 3.02 | 2.73 | 4.16 | 4.74 | 6.95 | 7.87 |
| C | * | * | 8.32 | 7.94 | 8.75 | 11.66 | 15.29 | 17.11 |

The accuracy of the proposed spike detection features combined with fine-grain features for cost prediction was significantly higher than the baseline features, the fine-grain features and the coarse-grain features (90.03 versus 86.99, 85.45 and 77.85; p<0.01). Similar significant differences were found in terms of recall and precision (Table 10) as well as Penalty Error (Table 11).

Table 10 - Accuracy, recall and precision of the proposed spike detection features combined with fine-grain features, the baseline features, the fine-grain features and the coarse-grain-features for cost prediction.

| Ftr | Acc | Recall | | | | | Precision | | | | |
|---|---|---|---|---|---|---|---|---|---|---|---|
| | | 1 | 2 | 3 | 4 | 5 | 1 | 2 | 3 | 4 | 5 |
| FS | 90.03 | 94.9 | 66.8 | 57.5 | 50.4 | 48.2 | 95.2 | 67.1 | 58.7 | 51.3 | 49.7 |
| C/F | 86.99 | 92.1 | 62.8 | 52.6 | 45.4 | 42.2 | 92.7 | 64.1 | 53.8 | 52.6 | 45.4 |
| F | 85.45 | 90.8 | 60.5 | 49.7 | 39.9 | 35.2 | 89.7 | 60.1 | 52.4 | 40.1 | 38.9 |
| C | 77.85 | 83.2 | 54.1 | 40.4 | 31.7 | 24.2 | 82.1 | 51.2 | 38.7 | 30.2 | 21.4 |

Acc is an abbreviation of Accuracy and Ftr is an abbreviation of Feature

Table 11 – Penalty Error of the proposed spike detection features combined with fine-grain features, the baseline features, the fine-grain features and the coarse-grain-features for cost prediction over the five cost buckets.

| Feature | All | 1 | 2 | 3 | 4 | 5 |
|---|---|---|---|---|---|---|
| FS | 0.23 | 0.17 | 0.54 | 0.59 | 0.66 | 0.70 |
| C/F | 0.32 | 0.22 | 0.83 | 0.97 | 1.08 | 1.11 |
| F | 0.65 | 0.57 | 0.99 | 1.23 | 1.45 | 1.61 |
| C | 1.02 | 0.89 | 1.62 | 1.89 | 2.06 | 2.29 |

### 5.3.4 Contribution of the proposed spike detection features

The MAPE of the combination of the proposed spike detection features for cost prediction was significantly higher than the MAPE of predictive models that did not include *Count_of_spike*, *Amount_of_positive_changes* and *Amount_of_negative_changes* (2.02 versus 2.41, 2.27 and 2.14; $p<0.01$) in their features (Table 12).

Table 12 - Effect of the proposed spike detection features on the cost prediction performance.

| Count_of_spike | Amount_of_positive_changes | Amount_of_negative_changes | All | 1 | 2 | 3 | 4 | 5 |
|---|---|---|---|---|---|---|---|---|
| * | * | * | **2.02** | **1.94** | **2.25** | **2.53** | **3.07** | **4.21** |
|   | * | * | 2.41 | 2.27 | 2.83 | 3.21 | 4.28 | 6.08 |
| * |   | * | 2.27 | 2.15 | 2.63 | 2.94 | 3.88 | 5.59 |
| * | * |   | 2.14 | 2.04 | 2.44 | 2.75 | 3.61 | 5.23 |

## 6. DISCUSSION

In this study, we investigated a method for leveraging patients' temporal data to predict healthcare costs. In doing so, this study makes three main contributions. First, we assessed the relative effect of different input features including cost, medical and visit information. We found no study in the literature that has established the relative effect of these three input feature types for future cost prediction. Second, we showed the deficiency of coarse-grain abstraction, which is the most common approach used in the literature to represent temporal data, and the superiority of fine-grain abstraction. Third, we demonstrated the high performance of the extracted temporal patterns for predicting patients' costs. This study suggests using fine-grain features rather than coarse-grain feature for enhancing the performance of cost prediction by detecting the patterns in the patients' temporal data.

We conducted one experiment to evaluate the effect of visits, medical and cost predictors, two experiments to test two hypotheses and one experiment to show the contribution of each proposed spike detection feature. The first experiment showed that cost predictors are the most important type of inputs for future cost prediction. Moreover, aligned with a previous study[7], adding medical features did not have any positive effect on the prediction performance. This showed that a patient's medical situation is likely implicitly represented by the cost predictors. In other words, adding medical features did not provide any further improvement. It should be mentioned that the medical features (i.e., drug, diagnosis and procedure groups) used in this study were limited to those proposed in the literature. There may still be different medical, drug, diagnosis and procedure groupings that might have more remarkable prediction power. However, adding visit features improved the prediction performance. This confirms H1b which hypothesizes that two patients with the same amount of cost in a specific time period may have dissimilar visit patterns that have differential impact on their future costs.

The second experiment aimed at comparing the effect of fine-grain and coarse-grain temporal abstraction. The fine-grain features significantly outperformed coarse-grain features. This confirmed our H1 which states that more detailed values from patients' monthly data can help improve prediction performance, and coarse-grain abstraction omits potentially useful data.

The third experiment showed the effect of proposed spike detection features to capture the patterns in patients' temporal data. This result retained our H2, since the proposed features combined with fine-grain features significantly outperformed all other methods. This superiority in numeric cost prediction accuracy was confirmed by the improved nominal prediction accuracy as well as by the domain knowledge based measure of the model that classifies a patient's future cost into one of the five cost buckets. It should be noted that while for privacy reasons we were not able to report the numeric prediction accuracy in terms of dollar values, the actual amount of correct prediction was very significant.

For example, after including all proposed features if an insurance company's payment for patients' medical expenses was $100 million, then the reduction in errors was about 28 million dollars. Compared to the strongest cost prediction method in the literature, Bertsimas et al. (2008)[7], the proposed features contributed to 25% error reduction on low cost patients and 54% error reduction on high cost patients. Moreover, this result showed that the proposed method had more effect (i.e., higher rate of error reduction) on the patients in the high cost bucket compared to the low cost bucket. This is specifically important since the high cost bucket is responsible for the majority of healthcare costs, although it includes only a limited number of patients [1, 3, 6]. However, similar to other literature [2, 7, 8, 10] on cost prediction, the performance for the high cost bucket was not as good as that for the low cost bucket. Finally, the fourth experiment showed that all three proposed spike detection features had a significant contribution on the superiority of the final prediction model.

The main limitation of this study was that we were only able to internally validate our method, and we did not apply it on another dataset to externally validate it as well. Thus, further validation of our approach with other datasets is needed. Moreover, another potential future direction is to study the effect of deep learning models for cost prediction, since they have shown to be very effective for feature learning. Finally, while we used the total amount of the paid cost by the health insurer as the target variable, breaking down the cost to a lower level (e.g., hospital or emergency) could help to better explore patients' costs. This could be a potential area for future research.

## 7. CONCLUSION

In this paper we attempted to improve the performance of healthcare cost predication methods by leveraging the rich information presented by fine-grain temporal abstraction of cost and visit information and the spikes in these fine-grain features over time. Also, while our study confirmed that cost predictors are strong inputs for healthcare cost prediction, it also suggested using visit information to enhance the performance. The reduction in error compared to the best established methods in the literature demonstrated the effectiveness of the proposed method to extract temporal patterns from a patient time series of data.